%% file: main.tex
\documentclass[10pt,twocolumn,letterpaper]{article}

\usepackage[pagenumbers]{cvpr} % To force page numbers, e.g. for an arXiv version

\input{preamble}

\definecolor{cvprblue}{rgb}{0.21,0.49,0.74}
\usepackage[pagebackref,breaklinks,colorlinks,allcolors=cvprblue]{hyperref}

\def\paperID{2818} % *** Enter the Paper ID here
\def\confName{CVPR}
\def\confYear{2025}

\title{\vspace{-20mm}
\ours{}: Pose-Free Feed-Forward 3D Gaussian Splatting \\ from Variable-length Image Sequence 
\vspace{-4mm}
}

\author{
Zequn Chen \quad Jiezhi ``Stephen'' Yang \quad Heng Yang \\[1mm]
Harvard University\\[2mm]
{\small\url{https://computationalrobotics.seas.harvard.edu/PreF3R/}}
}

\begin{document}

\twocolumn[{
    \renewcommand\twocolumn[1][]{#1}
    \maketitle
    \centering
    \vspace{-0.6cm}
    \input{figure/teaser}
    \vspace{0.6cm}
}]

\input{sec/0_abstract}    
\input{sec/1_intro}
\input{sec/2_related}
\input{sec/3_method}

\input{sec/4_experiment}
\input{sec/5_conclusion}
% \clearpage
{
    \small
    \bibliographystyle{ieeenat_fullname}
    \bibliography{main}
}

\input{sec/X_suppl}

\end{document}

%% file: preamble.tex
\newcommand{\ours}{\textsc{PreF3R}}

\newcommand{\deemph}[1]{{\color{black!40}#1}}

\usepackage{textcomp}
\usepackage{multirow}
\usepackage{float}

\definecolor{t1}{HTML}{B95454}
\definecolor{t2}{HTML}{444B8E}

%% file: figure/teaser.tex
\vspace{-4mm}
\includegraphics[width= 0.95\linewidth]{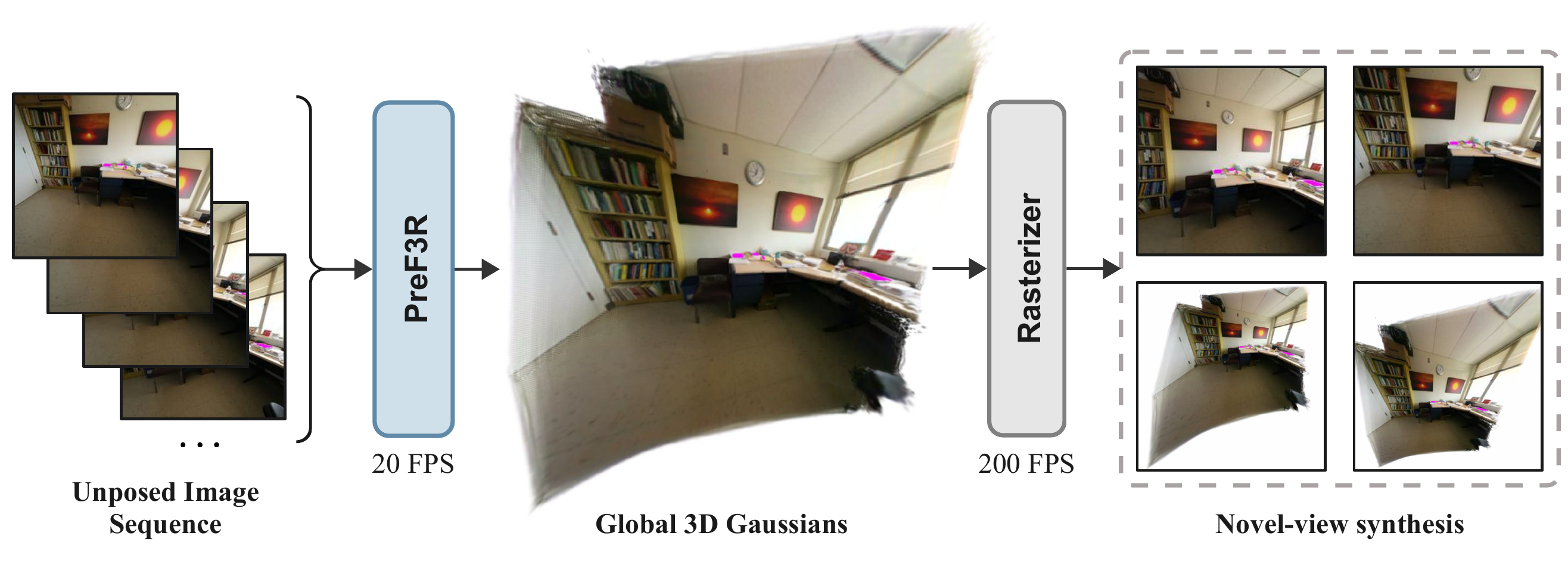}
\includegraphics[width= \linewidth]{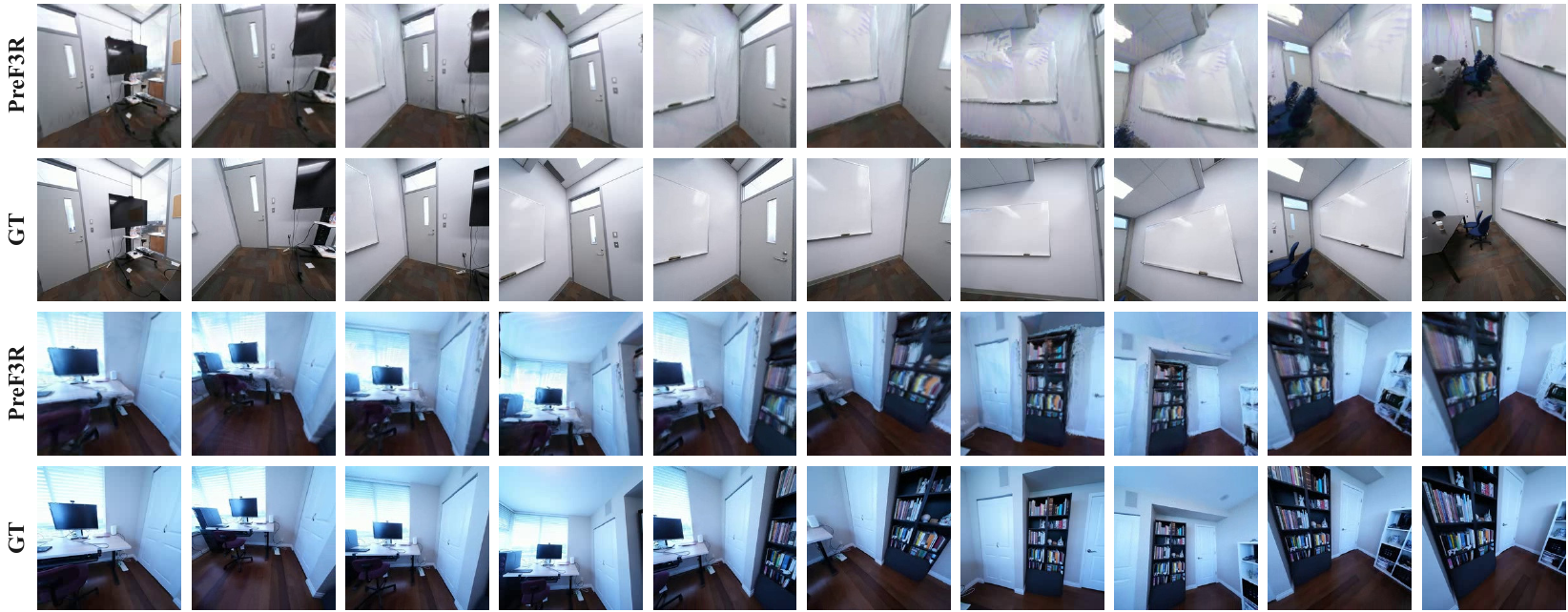}
\vspace{-6mm}
\captionof{figure}{
\textbf{Overview of \ours{}}. Given a sequence of unposed images of variable length, \ours{} incrementally reconstructs a set of 3D Gaussian primitives in a single feed-forward pass without any pre-processing or intermediate pose estimation. \ours{} operates at 20 FPS on a single H100 GPU, enabling real-time novel-view synthesis from numerous input images through differentiable rasterization.}

%% file: sec/0_abstract.tex
\begin{abstract}

\vspace{-4mm}
We present \ours{}, \underline{P}ose-F\underline{re}e \underline{F}eed-forward \underline{3}D \underline{R}econstruction from an image sequence of variable length. Unlike previous approaches, \ours{} removes the need for camera calibration and reconstructs the 3D Gaussian field within a canonical coordinate frame directly from a sequence of unposed images, enabling efficient novel-view rendering. We leverage DUSt3R's ability for pair-wise 3D structure reconstruction, and extend it to sequential multi-view input via a \emph{spatial memory network}, eliminating the need for optimization-based global alignment. Additionally, \ours{} incorporates a dense Gaussian parameter prediction head, which enables subsequent novel-view synthesis with differentiable rasterization. This allows supervising our model with the combination of photometric loss and pointmap regression loss, enhancing both photorealism and structural accuracy. Given a sequence of ordered images, \ours{} incrementally reconstructs the 3D Gaussian field at 20 FPS, therefore enabling real-time novel-view rendering. Empirical experiments demonstrate that \ours{} is an effective solution for the challenging task of pose-free feed-forward novel-view synthesis, while also exhibiting robust generalization to unseen scenes.

\end{abstract}

%% file: sec/1_intro.tex
\section{Introduction}
\label{sec:intro}

Rapid reconstruction of 3D scenes and synthesis of novel views from an unposed image sequence remains a challenging, long-standing problem in computer vision. Addressing this issue requires models that can simultaneously comprehend underlying camera distributions, 3D structures, color, and viewpoint information. Humans utilize this ability to form mental 3D representations for spatial reasoning, a capability that is also crucial across applications in 3D content creation, augmented/virtual reality, and robotics, etc.

Traditional approaches for novel-view synthesis often require posed images as inputs, or would first estimate 3D structure through techniques such as Structure-from-Motion (SfM)~\cite{crandall2011discrete,wilson2014robust,sweeney2015optimizing,snavely2006photo,agarwal2009building,wu2013towards,schonberger2016colmap}, Bundle Adjustment~\cite{triggs2000bundle,agarwal2010bundle,wu2011multicore,yu2024sim} and/or SLAM~\cite{davison2007monoslam,klein2007parallel,newcombe2011dtam}. These techniques typically rely on hand-crafted heuristics~\cite{pan2024globalstructurefrommotionrevisited}, can be complex and slow to optimize, and may exhibit instability in challenging scenes. Once the camera poses are estimated, differentiable rendering methods like NeRF, Instant-NGP, and Gaussian Splatting~\cite{mildenhall2020nerf, mueller2022instant, kerbl20233dgaussiansplattingrealtime} can be applied for novel-view synthesis via another per-scene optimization with photometric loss, which further slows down the overall pipeline. Moreover, the separate multi-stage solutions can potentially lead to suboptimal results~\cite{hong2023unifying}.

Other research approached novel-view synthesis and 3D reconstruction by jointly optimizing these tasks within a unified learning-based framework~\cite{lin2021barfbundleadjustingneuralradiance, xu2024sparpfast3dobject, wang2022nerfneuralradiancefields, chen2023dbarfdeepbundleadjustinggeneralizable}. While these approaches generally allow for pose-free 3D reconstruction and novel-view synthesis, some of them still require pre-computed coarse camera poses. Besides, the interdependence of these tasks may lead to the ``chicken-and-egg'' problem, impairing the rendering quality compared to methods with known poses~\cite{lin2021barfbundleadjustingneuralradiance}. These methods also necessitate per-scene optimization, which is costly and sensitive due to the nonconvexity of the optimization~\cite{chen2023dbarfdeepbundleadjustinggeneralizable}.

Recent advancements favor prior-based reconstruction approaches that leverage generalizable learned priors~\cite{detone2018superpoint,sarlin2020superglue,yao2018mvsnet,bloesch2018codeslam,sun2021loftr,yin2023metric3d,wang2024vggsfm,he2024dfsfm}. Among them, some notable models like DUSt3R~\cite{wang2024DUSt3R} and MASt3R~\cite{leroy2024groundingimagematching3d} propose to regress pointmaps from uncalibrated \emph{image pairs} with a transformer-based architecture trained on large-scale 3D datasets. Multiview inputs, however, would still require further optimization based on explicit pose estimation. Follow-up work, such as Spann3R~\cite{wang20243dreconstructionspatialmemory}, introduces the spatial memory network that extends the model to project pointmaps from multiple views into a canonical coordinate frame, eliminating the need for optimization-based global alignment. A very near-term line of works ~\cite{ye2024poseproblemsurprisinglysimple, smart2024splatt3rzeroshotgaussiansplatting, fan2024largespatialmodelendtoend} extend the power of pretrained models like DUSt3R and MASt3R to 3D Gaussian reconstruction by incorporating a Gaussian prediction head, and achieves remarkable performance in pose-free novel-view synthesis task. However, due to reliance on DUSt3R and MASt3R, these methods inherit the limitation of pairwise inputs, constraining their scalability.

\paragraph{Contribution.} We present \ours{}, 
% which to the best of our knowledge, is 
the first \emph{pose-free}, \emph{feed-forward} framework for online 3D Gaussian reconstruction from a variable-length image sequence. 
To mitigate the requirement for camera poses, \ours{} leverages the capability of the pretrained reconstruction model DUSt3R for pairwise 3D structure prediction, and extends it to multi-view inputs via a \emph{spatial memory network}~\cite{wang20243dreconstructionspatialmemory}. This, crucially, enables sequential projection of pointmaps from multiview images into the canonical 3D space (see Fig.~\ref{fig:method}).
DUSt3R's ViT-based~\cite{dosovitskiy2020image} encoder-decoder structure and dense-prediction head (DPT)~\cite{ranftl2021vision} make it convenient for \ours{} to further integrate Gaussian parameter prediction for each 3D point. Specifically, the decoder outputs each view's pointmap in the canonical space, interpreted as the center of Gaussian points, with an extra head that estimates the corresponding Gaussian parameters. This facilitates fast novel-view synthesis through differentiable rasterization and allows us to supervise the model with a combination of photometric loss and point regression loss during training, promoting a high level of photorealism and structural accuracy.
During inference, given an ordered image sequence of unlimited length, \ours{} reconstructs 3D Gaussian in a canonical space for the \emph{entire scene} in a purely feed-forward manner. \ours{} operates at 20 FPS on a single Nvidia H100 GPU, demonstrating high-quality rendering results and strong generalization capabilities.

% \hank{Should add brief description of the experimental results here.}
\vspace{-2mm}
\paragraph{Paper organization.}
% \hank{describe paper structure here briefly}
The remainder of this paper is organized as follows: 
In Sec.~\ref{sec:related}, we review related works pertinent to our research, summarizing recent advancements and highlighting gaps that our work addresses. Sec.~\ref{sec:method} details our proposed methodology, including the underlying  framework and novel algorithmic components. In Sec.~\ref{sec:experiment}, we describe the experimental setup and present results with in-depth analysis. We conclude in Sec.~\ref{sec:conclusion} and comment on limitations and future research directions. 

%% file: sec/2_related.tex
\section{Related Works}
\label{sec:related}

\subsection{Two-stage 3D reconstruction and NVS}

Traditional 3D reconstruction and novel-view synthesis can be split into two stages. The first stage involves estimating camera parameters using geometric principles and multi-view stereo techniques. These include Structure from Motion (SfM)~\cite{crandall2011discrete,wilson2014robust,sweeney2015optimizing,snavely2006photo,agarwal2009building,wu2013towards,schonberger2016colmap} and stereo matching~\cite{furukawa2009furu,schonberger2016pixelwise,galliani2015massively}. SfM constructs sparse 3D point clouds by aligning images based on feature correspondences, typically followed by bundle adjustment~\cite{triggs2000bundle,agarwal2010bundle,wu2011multicore,yu2024sim}. Techniques like Visual Simultaneous Localization and Mapping (V-SLAM)~\cite{davison2007monoslam,klein2007parallel,newcombe2011dtam} further enrich this process through real-time integration of motion estimation and mapping. However, these methods often encounter efficiency challenges, particularly with respect to computational complexity of the hand-crafted features and underlying optimization~\cite{pan2024globalstructurefrommotionrevisited}. The iterative nature of feature matching, along with the high dimensionality of data and need for search and optimization algorithms, can result in substantial computational overhead.

To synthesize novel viewpoints, differentiable rendering methods like NeRF~\cite{mildenhall2020nerf} and its numerous extensions~\cite{barron2022mip,kerbl2023gaussian,barron2023zip,wang2021neus,yariv2021volsdf,jang2021codenerf,jang2024nvist,huang20242d, yang2024carffconditionalautoencodedradiance} facilitate high-fidelity image generation by optimizing models on images with known camera parameters. While recent advancements have significantly improved the speed of neural rendering--- for example, Gaussian Splatting~\cite{kerbl2023gaussian} achieves rendering frame rates over 100fps--- these approaches still require extensive per-scene optimization time at test time, often spending minutes to fit the model prior to rendering.

\subsection{Joint optimization}

To avoid the need of separately running a camera calibration pipeline, various work seeks to co-optimize reconstruction with differentiable rendering. Specifically, NeRF~\cite{wang2022nerfneuralradiancefields} jointly optimizes NeRF parameters and camera pose embeddings during training through a photometric reconstruction, while SiNeRF~\cite{xia2022sinerfsinusoidalneuralradiance} further improves optimality in NeRF. BARF~\cite{chen2023dbarfdeepbundleadjustinggeneralizable} and its extension GARF~\cite{chng2022garfgaussianactivatedradiance} employ a coarse-to-fine strategy, learning low-frequency components before gradually registering high-frequency details, thus addressing issues from imprecise camera poses. To extend it to multiple scales, RM-NeRF~\cite{jain2022robustifyingmultiscalerepresentationneural} combines a GNN-based motion averaging network with Mip-NeRF~\cite{barron2021mipnerfmultiscalerepresentationantialiasing}, while GNeRF~\cite{meng2021gnerfganbasedneuralradiance} uses adversarial learning to estimate camera poses through a generator-discriminator setup.

These methods, however, are limited in their generalizability and demand a sensitive, time-intensive per-scene optimization. Additionally, several of them rely on coarse camera pose initialization provided by SfM~\cite{lin2021barfbundleadjustingneuralradiance}. To address the high computational costs of the above approaches, DBARF~\cite{chen2023dbarfdeepbundleadjustinggeneralizable} introduces a multi-stage process to learn a cost feature map, while InstantSplat~\cite{fan2024instantsplatsparseviewsfmfreegaussian} utilizes feed-forward models to facilitate structure reconstruction. Although these methods reduce computational expense, they still require some levels of joint optimization to ensure multi-view consistency, and fall short of enabling a fully feed-forward pipeline for 3D reconstruction and novel-view synthesis.

\subsection{Feed-forward novel-view synthesis}

Recently, learning-based, feed-forward Gaussian Splatting methods have been proposed. By pretraining the models on vast amounts of 3D data, these techniques aim to directly infer novel views from 2D images in a feed-forward manner. Notably, Splatter Image~\cite{szymanowicz2024splatterimageultrafastsingleview} regresses pixel-aligned Gaussians directly from a single image input, and pixelSplat~\cite{charatan2024pixelsplat3dgaussiansplats} extends the idea to paired input, but is still subject to noisy geometry reconstruction. MVSplat~\cite{chen2024mvsplat}, on the other hand, works with a flexible-length input image sequence, but requires image poses as inputs.

Feed-forward techniques to recover accurate 3D structure is another line of research. These works include monocular depth estimation~\cite{dexheimer2023learning,yin2023metric3d,ke2024repurposing}, multi-view depth estimation~\cite{yao2018mvsnet,duzceker2021deepvideomvs,sayed2022simplerecon}, optical flow~\cite{teed2020raft}, point tracking~\cite{doersch2023tapir,karaev2023cotracker,xiao2024spatialtracker}, etc. DUSt3R~\cite{wang2024DUSt3R} offers a unified 3D reconstruction approach by directly learning to map an image pair to 3D pointmap, followed by an optimization-based global alignment to project pointmaps into a canonical coordinate system. MASt3R~\cite{leroy2024groundingimagematching3d} additionally learns dense local features on top of DUSt3R~\cite{wang2024DUSt3R} for more robust 3D feature matching. 
These methods serve as a strong subroutine for various Gaussian Splatting reconstruction methods from stereo inputs~\cite{ye2024poseproblemsurprisinglysimple, smart2024splatt3rzeroshotgaussiansplatting, fan2024largespatialmodelendtoend}, which predict Gaussian parameters on top of a DUSt3R~\cite{wang2024DUSt3R}-based method.
To eliminate the necessity of the global alignment step in DUSt3R\cite{wang2024DUSt3R}, Spann3R~\cite{wang20243dreconstructionspatialmemory} employs an external spatial memory to progressively reconstruct 3D structures. This memory system learns to track all previously acquired relevant 3D information and enables incrementally projecting new images into a canonical space. Our work shares a similar spirit as Spann3R but takes a significant step forward: we leverage spatial memory to handle more than two views but also predict 3D Gaussians and enable novel-view synthesis.

% However, it's not applicable to novel-view rendering.

In summary, developing a truly pose-free, feed-forward, and online pipeline for 3D reconstruction and novel-view synthesis has remained a significant challenge. To the best of our knowledge, \ours{} represents the first model capable of achieving this feature, promoting potential breakthroughs in applications including augmented and virtual reality, robotics, self-driving, and beyond.

%% file: sec/3_method.tex
\section{Method}
\label{sec:method}

\begin{figure*}
    \centering
    \includegraphics[width= \textwidth]{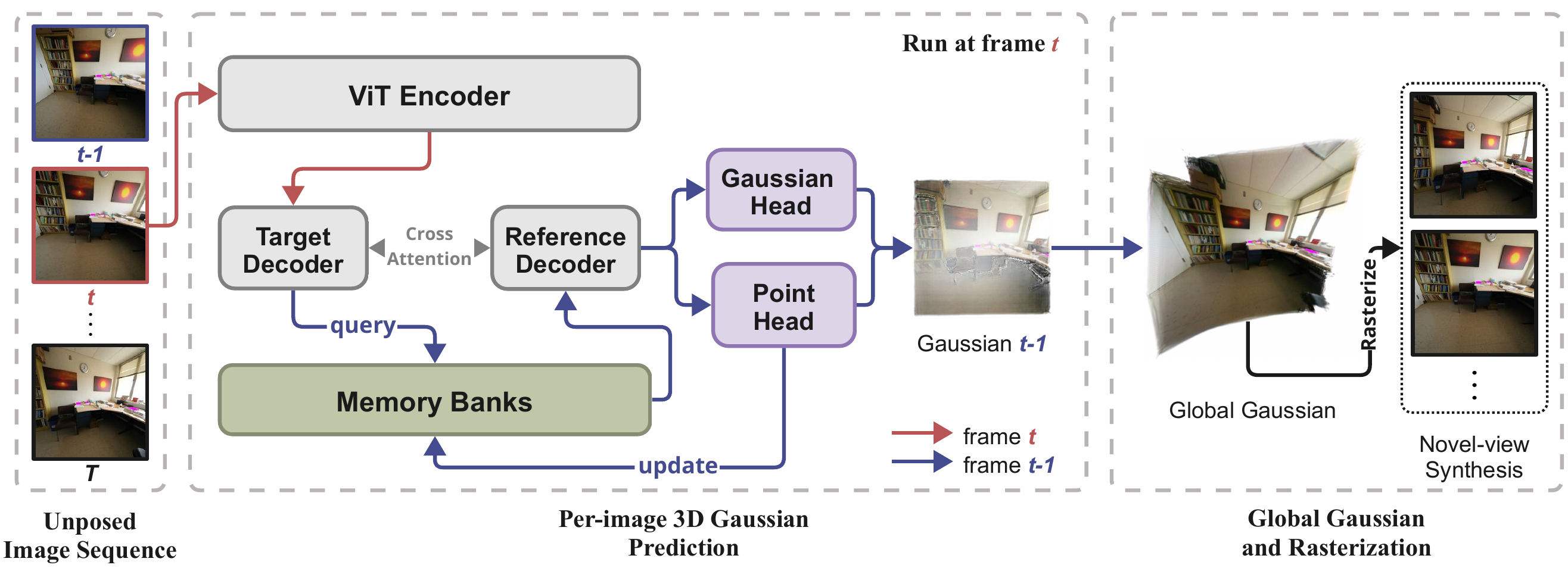}
    \caption{\textbf{\ours{}'s overall architecture}. \textbf{Left:} An ordered set of unposed images $\{I_t\}_{t=1}^T$ is fed into \ours{} sequentially. \textbf{Middle:} At timestamp $t$, the input frame \textcolor{t1}{$I_t$} is first encoded by a ViT-encoder into \textcolor{t1}{$f_t$}, which is then decoded into the query feature \textcolor{t1}{$f_t^q$} by the Target Decoder. The Target Decoder is intertwined with the Reference Decoder through cross-attention. Simultaneously, the query feature of the previous frame \textcolor{t2}{$f_{t-1}^q$} queries the memory bank to produce the fused feature \textcolor{t2}{$f_{t-1}^g$}, which the Reference Decoder decodes into the output feature \textcolor{t2}{$f_{t-1}^h$}. \textcolor{t2}{$f_{t-1}^h$} is then processed by the Gaussian Head and the Point Head to produce pixel-aligned Gaussian primitives. \textbf{Right:} The output from each frame is accumulated into global Gaussian primitives, enabling fast novel-view synthesis through rasterization.}
    \label{fig:method}
\end{figure*}

\subsection{Problem formulation}

The objective of our work is to (a) reconstruct a 3D scene and (b) achieve real-time novel-view synthesis from a sequence of \( T \) unposed image frames, denoted as \( \{ I_t \}_{t=1}^{T} \), where each image \( I_t \in \mathbb{R}^{H \times W \times 3} \) represents RGB pixel values of resolution \( H \times W \). Specifically, given this sequence of unposed images, our goal is to develop a model capable of reconstructing a set of 3D Gaussian primitives for the entire scene in a single feed-forward pass, and enable fast synthesis of photorealistic renderings \( \hat{I}_t \) from novel viewpoints through a differentiable rasterization process.

The challenging here is to develop a non-optimization-based model capable of both 3D reconstruction and novel-view synthesis from image sequences alone. The simultaneous pursuit of handling pose-free input and feed-forward architecture significantly amplifies the complexity, and necessitates robust strategies for capturing the essential spatial structure. Furthermore, employing a feed-forward model imposes a key constraint that each pass through the network must efficiently extract and utilize spatial information without iterative refinement. 

With these goals and constraints in mind, we present the overall model architecture of \ours{} in Fig.~\ref{fig:method}. In the next sections, we provide a detailed description of our model.

\subsection{Structural reconstruction}
\label{subsec:structure}

For 3D structural reconstruction, we adopt a ViT-based model equipped with a dense-prediction (DPT) head similar to those used in DUSt3R~\cite{wang2024DUSt3R} and MASt3R~\cite{leroy2024groundingimagematching3d}. This setup enables generalizable feed-forward 3D reconstruction from stereo images, supervised by pointcloud regression loss and confidence loss. However, the key limitation of these models is that they struggle in handling more than two images, as they would necessitate an optimization-based global alignment process to establish spatial correspondence among multiple views. To address this limitation, we further incorporate a spatial memory network ~\cite{wang20243dreconstructionspatialmemory} to extend our approach to multi-view inputs. 

Given an ordered image sequence without poses \( \{ I_t \}_{t=1}^{T} \), we project the pointmaps from all views into a canonical space, specifically, the coordinate system of the first input view.  Initially, a ViT-based encoder with shared weights encodes the images into features \( f_t = \mathrm{Enc}(I_t)\). For each incoming feature \( f_t \), the query feature from the previous frame \( f_{t-1}^q \) is used to query the key and value features \( (f^k, f^v) \) from the memory, thus generating a fused feature \( f_{t-1}^g \). Following this, the image feature and fused feature \( (f_t, f_{t-1}^g) \) are processed by the ViT decoder equipped with cross-attention to produce space-aware features \( (f_{t}^{h'}, f_{t-1}^h) \). The resulting feature \( f_{t}^{h'} \) is input into the query MLP head $\mathrm{MLP}_\text{query}$ to generate the query feature \( f_t^q \) for subsequent processing:
\begin{equation}
f_t^q = \mathrm{MLP}_\text{query}(f_{t}^{h'}, f_{t}).
\end{equation}
The other output feature \( f_{t-1}^h \) is fed into the output MLP head to output the pointmap and confidence:
\begin{equation}
\hat{X}_{t-1}, C_{t-1} = \mathrm{MLP}_\text{out}(f_{t-1}^h).
\end{equation}

To conserve GPU memory, the memory banks are segmented into working memory and long-term memory, enabling efficient management of resources for numerous input views as in ~\cite{wang20243dreconstructionspatialmemory}. The working memory retains complete memory features for the most recent \( N_\text{working} \) frames. When this memory reaches capacity, the oldest features are transferred to the long-term memory. In the long-term memory, each token's accumulated attention weights are monitored, and upon surpassing a predefined threshold, only the top-$k$ tokens are retained for sparsification.

\subsection{3D Gaussian Splatting}

\paragraph{Definition.}
3D Gaussian Splatting (3D-GS)~\cite{kerbl2023gaussian} defines a radiance field by assigning an opacity \( \sigma(x) \in \mathbb{R}^+ \) and a color \( c(x, \nu) \in \mathbb{R}^3 \) to each 3D point \( x \in \mathbb{R}^3 \) and viewing direction \( \nu \in \mathbb{S}^2\), where \( \sigma \) and \( c \) are represented by a mixture \( \theta \) of \( G \) colored 3D Gaussian primitives:
\begin{equation}
g_i(x) = \exp \left( -\frac{1}{2} (x - \mu_i)^\top \Sigma_i^{-1} (x - \mu_i) \right),
\end{equation}
where \( \mu_i \in \mathbb{R}^3 \) is the mean and \( \Sigma_i \in \mathbb{R}^{3 \times 3} \) is the covariance, defining the shape and size of the Gaussian. Each Gaussian has an opacity \( \sigma_i \in [0, 1] \) and a view-dependent color \( c_i(\nu) \in \mathbb{R}^3 \), which, together, define:
\begin{equation}
\sigma(x) = \sum_{i=1}^G \sigma_i g_i(x), \ c(x, \nu) = \frac{\sum_{i=1}^G c_i(\nu) \sigma_i g_i(x)}{\sum_{j=1}^G \sigma_j g_j(x)}.
\end{equation}

3D-GS~\cite{kerbl2023gaussian} provides an efficient, differentiable renderer \( \hat{I} = R(\theta, \pi) \), mapping the Gaussian primitives \( \theta = \{ (\sigma_i, \mu_i, \Sigma_i, c_i), \; i = 1, \dots, G \} \) and camera view $\pi \in \mathbb{SE}(3)$ to the output image \( \hat{I} \).

\paragraph{Traditional approaches.}
The original 3D-GS uses an iterative process to fit Gaussian splats to a single scene. However, Gaussian primitives experience vanishing gradients when distant from their target location~\cite{charatan2024pixelsplat3dgaussiansplats}. To mitigate this, 3D-GS initializes with SfM point clouds and applies non-differentiable ``adaptive density control" for splitting and pruning Gaussians~\cite{kerbl2023gaussian}. While effective, this approach requires dense image collections and does not support generalizable, feed-forward models that predict Gaussians without per-scene optimization.

\paragraph{Feed-forward Gaussian Splatting.}
Recent models such as pixelSplat and MVSplat~\cite{charatan2024pixelsplat3dgaussiansplats, chen2024mvsplat} utilize a dense-prediction head~\cite{ranftl2021vision} to directly predict pixel-aligned 3D Gaussian parameters from multi-view inputs. \ours' structural reconstruction component also incorporates a dense-prediction head, which generates per-pixel pointmaps and confidence maps. It is intuitive to implement an additional Gaussian MLP head following the decoder, in parallel with \( \mathrm{MLP}_\text{out} \), to produce pixel-aligned Gaussian primitives for each frame:
\begin{equation}
\theta_{t} = \{ (\sigma_i, \mu_i, \Sigma_i, c_i) \}_{i=1}^{H \times W} = \mathrm{MLP}_\text{GS}(f_{t}^h),
\end{equation}
we use Spherical Harmonics (SH) to represent $c_i \in \mathbb{R}^{3\times d} $, where $d$ is the degree of SH. After this, we employ a fast differentiable rasterizer to render images for targeted views from the predicted Gaussian primitives. 

With this architecture, our model can simultaneously generate both pointmaps and rendered images for novel-view synthesis in a \textit{single feed-forward} pass. Importantly, this dual output facilitates joint supervision through confidence-based pointmap regression loss and photometric loss, as described next.

\input{table/benchmark_Scannetpp}
\input{table/benchmark_ArkitScenes}

\subsection{Training and inference}
\label{subsec:training}
\paragraph{Training.}
For pointmap regression, we use the same loss function as in DUSt3R~\cite{wang2024DUSt3R}, extending its formulation from pairwise to multiview inputs.:
% \begin{equation}
\begin{gather}
    \mathcal{L}_\text{conf} = \sum_{t=1}^{T} \sum_{i}^{H \times W} C_{t}^{i} \ell_\text{regr}(t, i) + \alpha \log (C_{t}^{i}), \\
    \ell_\text{regr}(t, i) = \left\lVert \frac{1}{z} X_t^i -  \frac{1}{z} \hat{X}_t^i \right\rVert,
\end{gather}
% \end{equation}
where $\alpha$ is a regularization hyperparameter, and $z$ represents the estimated scale factor used to resolve scale ambiguity between the predicted and ground-truth pointmaps. The scale factor is calculated explicitly from the norm of the estimated global pointmaps.

\begin{equation}
    \mathcal{L}_\text{MMSE} = \mathcal{L}_\text{MSE} (M \cdot I_t, M \cdot \hat{I}_t),
\end{equation}
where the mask $M$ is defined as setting the region where predicted alpha is less than a threshold $th_\text{alpha}$ to zero. The overall loss is formulated as a weighted combination of these two losses:
\begin{equation}
    \mathcal{L} = \mathcal{L}_\text{conf} + \lambda \mathcal{L}_\text{MMSE}.
\end{equation}

\begin{figure}
    \centering
    \includegraphics[width=1\linewidth]{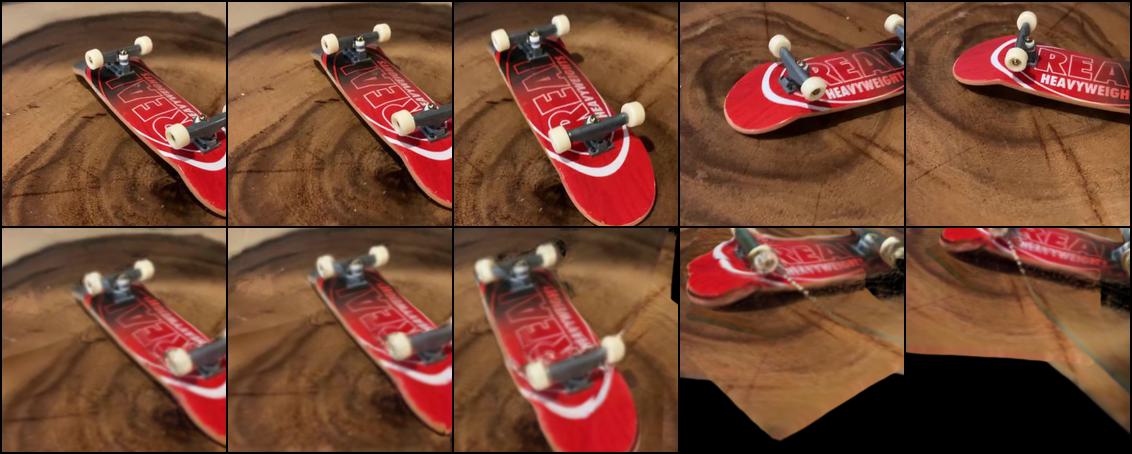}
    \caption{\textbf{Scale ambiguity problem.} Even slight scale shifts can cause significant view drifts in rendered results from ground-truth camera poses, making it hard to apply photometric supervision. \textbf{Top row}: ground-truth images; \textbf{Bottom row}: rendered images. Data sample is from Co3D~\cite{reizenstein2021common}.}
    \label{fig:scale_ambiguity}
\end{figure}

Although \ours's DPT head allows for the direct prediction of pointmaps (i.e., \( \mu \) of gaussian primitives) from multi-view inputs, the issue of scale ambiguity between the predicted and ground-truth pointmaps and camera translation still remains. This ambiguity is detrimental when rendering images using ground-truth camera poses for photometric supervision. As shown in Fig.~\ref{fig:scale_ambiguity}, even a minor error in scale factor estimation can lead to significant view drift in the rendered images. We observe that the estimated scene scale factor 
$z$ performs effectively only for specific datasets, affected by the complex natural conditions of the captured scenes and camera perspectives. Therefore, to prevent training-time view drifts during rendering, we choose datasets where the scale factor is sufficiently well estimated to train our model; see details in Sec.~\ref{subsec:implement}. 

To save GPU memory during training, we fix the number of input views to  $ N_\text{train} $. Given the high sensitivity of the rendering process to the accuracy of the predicted structure, we avoid training the model on image sequences with large baselines or minimal frame overlap. This approach contrasts with prior methods~\cite{keetha2024splatam, wang20243dreconstructionspatialmemory}.

Additionally, to enhance rendering quality, we propose incorporating additional target views for photometric supervision during training, similar to previous works~\cite{keetha2024splatam, charatan2024pixelsplat3dgaussiansplats, chen2024mvsplat}. Specifically, we sample $N_\text{extra}$ frames between each pair of input frames, and apply differentiable rasterization and calculate MSE loss for these selected views. We set a strategy to sample within $( T_{\min}, T_{\max} )$ interval for adjacent frames. 

% \hank{For these extra views, we do not have groundtruth images right? Where does supervision come from?} \zequn{Yes we do have the ground-truth images as well as camera poses, as they were sampled from original data frames, not by interpolating camera views.}

\paragraph{Inference.}
During inference, \ours{} maintains a long-term memory to track spatial relationships across multiple inputs, as described in Section~\ref{subsec:structure}. For better rendering quality, we prune the predicted Gaussian primitives whose confidence is lower than the preset threshold $th_{\text{conf}}$. Unlike the training phase, as ground-truth camera poses and pointmaps are unavailable, the global Gaussian primitives $\theta_\text{global} = \bigcup_{t=1}^{T} \theta_{t} $ can be saved and visualized using offline rendering tools~\cite{kerbl2023gaussian, ye2024gsplatopensourcelibrarygaussian}. Alternatively, the camera poses of input frames can be estimated through an extra optimization process as described in ~\cite{wang2024DUSt3R, leroy2024groundingimagematching3d, wang20243dreconstructionspatialmemory}, which is, however, beyond the scope of our work.

%% file: table/benchmark_Scannetpp.tex
\begin{table*}[ht]
\begin{center}
\setlength{\tabcolsep}{1.0mm}{

\begin{tabular}{lllcccccccccccc} 
\hline
\multirow{2}{*}{FF} & \multirow{2}{*}{PF} & \multirow{2}{*}{Method} & \multicolumn{4}{c}{2 views}                                       & \multicolumn{4}{c}{10 views}                                      & \multicolumn{4}{c}{50 views}                                       \\ 
\cline{4-15}
                    &                     &                         & PSNR$\uparrow$           & SSIM$\uparrow$           & LPIPS$\downarrow$         & time           & PSNR$\uparrow$          & SSIM$\uparrow$          & LPIPS$\downarrow$         & time           & PSNR$\uparrow$          & SSIM $\uparrow$          & LPIPS $\downarrow$         & time            \\ 
\hline
\deemph{\checkmark}          &                     & \deemph{MVSplat}                 &  \deemph{23.67}            & \deemph{0.811}         &    \deemph{0.180}      &   \deemph{0.081}             &   \deemph{19.31}             &  \deemph{0.697}              &    \deemph{0.334}            &      \deemph{0.554}           &      \deemph{-}          &      \deemph{-}          &     \deemph{-}           &       \deemph{-}          \\
                    & \deemph{\checkmark}          & \deemph{InstantSplat}            &    \deemph{20.91}       &     \deemph{0.766}      &  \deemph{0.362}         &     \deemph{49.75}      &     \deemph{17.32}      &  \deemph{0.623}         &   \deemph{0.389}             &      \deemph{92.49}         &    \deemph{16.89}            & \deemph{0.571}               &    \deemph{0.410}        &     \deemph{148.9}      \\
\hline
\checkmark          & \checkmark          & Spann3R                 &  22.36         &  0.788         &   0.128        &  \textbf{0.119} &   21.86        &     0.779      &     0.134     &     \textbf{0.451}           &   19.75        &     0.669      &    0.221            &    \textbf{2.039}             \\
\checkmark          & \checkmark          & Splatt3R                &  18.54         &  0.659         &   0.283        &  {0.137}        & -              & -              & -              & -              & -              & -              & -              & -               \\ 
\hline
\checkmark          & \checkmark          & \textbf{ours}           & \textbf{22.83} & \textbf{0.800} & \textbf{0.124} & {0.146} & \textbf{22.60} & \textbf{0.793} & \textbf{0.128} & {0.472} & \textbf{20.38} & \textbf{0.702} & \textbf{0.206} & {2.265}  \\
\hline
\end{tabular}

\caption{\textbf{Novel-view synthesis performance on Scannet++~\cite{yeshwanthliu2023scannetpp}}. All metrics are averaged over 10 validation scenes. For input views of 2, 10, and 50, the frame sampling intervals are 5, 3, and 2, respectively. Average running time is measured in seconds. ``FF'' denotes Feed-Forward, and ``PF'' denotes Pose-Free.  MVSplat~\cite{chen2024mvsplat} encounters CUDA out-of-memory on 50-view scenes on our H100 GPU. Spann3R~\cite{wang20243dreconstructionspatialmemory} is a pointmap prediction model, we evaluate its rendering results by projecting the predicted colored pointmaps back onto image planes. Splatt3R~\cite{keetha2024splatam} is a pose-free feed-forward model that only handles 2-view input. }
\label{tab:scannetpp}

}
\end{center}
\vspace{-0.3cm}
\end{table*}

%% file: table/benchmark_ArkitScenes.tex
\begin{table*}[ht]
\begin{center}
\setlength{\tabcolsep}{1.0mm}{

\begin{tabular}{lllcccccccccccc} 
\hline
\multirow{2}{*}{FF} & \multirow{2}{*}{PF} & \multirow{2}{*}{Method} & \multicolumn{4}{c}{2 views}                                       & \multicolumn{4}{c}{10 views}                                      & \multicolumn{4}{c}{50 views}                                       \\ 
\cline{4-15}
                    &                     &                         & PSNR$\uparrow$           & SSIM$\uparrow$           & LPIPS$\downarrow$         & time           & PSNR$\uparrow$          & SSIM$\uparrow$          & LPIPS$\downarrow$         & time           & PSNR$\uparrow$          & SSIM $\uparrow$          & LPIPS $\downarrow$         & time            \\ 
\hline
\deemph{\checkmark}          &                     & \deemph{MVSplat}                 &    \deemph{23.00}            &     \deemph{0.782}           &    \deemph{0.210}            &    \deemph{0.079}            &    \deemph{19.43}           &    \deemph{0.694}            &     \deemph{0.357}           &      \deemph{0.506}          &     \deemph{-}           &          \deemph{-}      &     \deemph{-}           &    \deemph{-}             \\
                    & \deemph{\checkmark}          & \deemph{InstantSplat}            &   \deemph{19.79}    &   \deemph{0.656}        &  \deemph{0.213}        &       \deemph{48.99}       &  \deemph{18.55}              &     \deemph{0.586}       &   \deemph{0.314}  &     \deemph{97.56}         &   \deemph{16.73}             &    \deemph{0.574}            &  \deemph{0.397}            &       \deemph{152.2}      \\
\hline
\checkmark          & \checkmark          & Spann3R                 &    19.50       &      0.648    &   \textbf{0.172}         &    \textbf{0.115}            &     19.96      &       0.631    &    0.218            &    \textbf{0.448}            &    17.56       &      0.505    &   0.335  & \textbf{1.919}                \\
\checkmark          & \checkmark          & Splatt3R                &    17.46       &      0.517    &   0.349  & {0.122} & -              & -              & -              & -              & -              & -              & -              & -               \\ 
\hline
\checkmark          & \checkmark          & \textbf{ours}           & \textbf{20.96} & \textbf{0.712} & {0.187} & {0.152}        & \textbf{21.91} & \textbf{0.693} & \textbf{0.210} & {0.490} & \textbf{18.70} & \textbf{0.563} & \textbf{0.300} & {2.232}  \\
\hline
\end{tabular}

\caption{\textbf{Novel-view synthesis performance on ArkitScenes~\cite{baruch1arkitscenes}}. All metrics are averaged over 10 validation scenes. Average running time is measured in seconds. Experimental settings and evaluation criteria are consistent with those described in Tab.~\ref{tab:scannetpp}. }
\label{tab:arkit}
}
\end{center}
\vspace{-0.3cm}
\end{table*}

%% file: sec/4_experiment.tex
\section{Experiment}
\label{sec:experiment}
\begin{figure*}
    \centering
    \includegraphics[width= \textwidth]{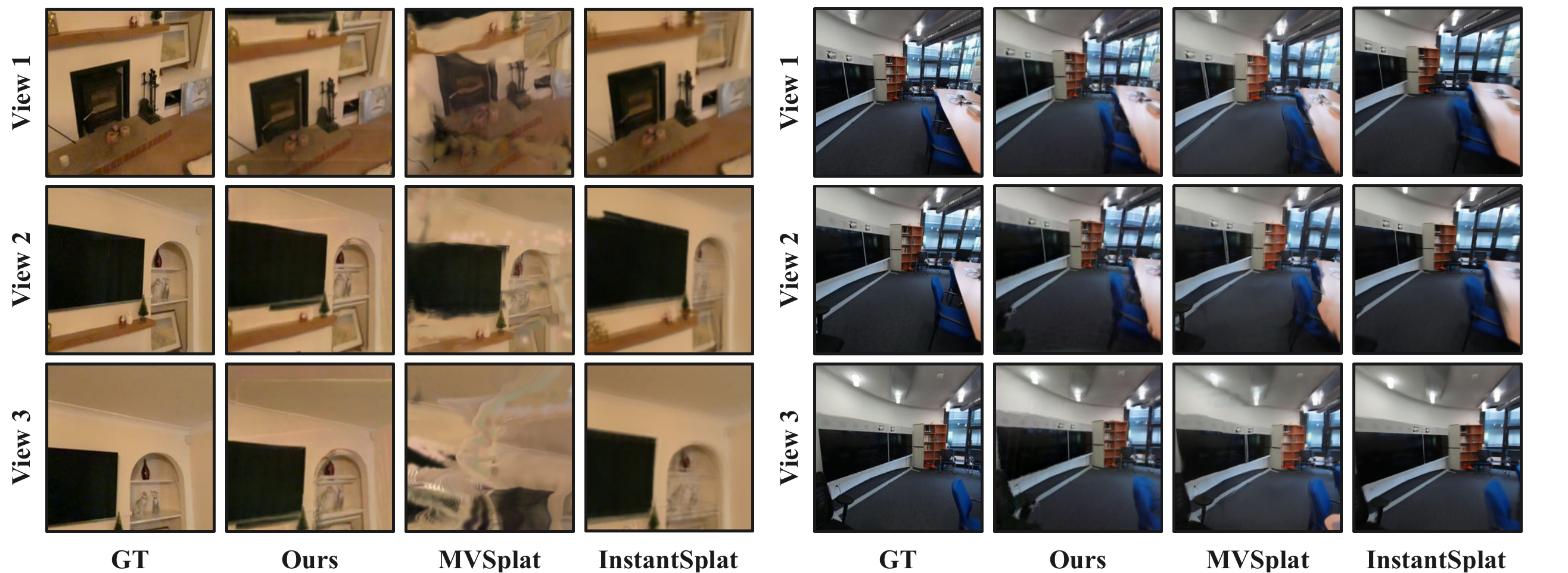}
    \caption{\textbf{Qualitative comparison of novel view synthesis performance.} \textbf{Left}: visualization of scene reconstructions from ARKitScenes~\cite{baruch1arkitscenes}; \textbf{Right}: visualization of reconstructions from ScanNet++~\cite{yeshwanthliu2023scannetpp}. Each row corresponds to a unique viewpoint, while each column displays the output of a different method. Note that MVSplat~\cite{chen2024mvsplat} relies on ground truth poses, and InstantSplat~\cite{fan2024instantsplatsparseviewsfmfreegaussian} requires per-scene optimization, whereas \ours{} requires neither. \ours{} achieves comparable or superior photorealism and demonstrates better structural accuracy relative to the other methods.}

    \label{fig:compare_viz}
\end{figure*}

\subsection{Implementation details}
\label{subsec:implement}

\paragraph{Dataset.}
As mentioned in Section~\ref{subsec:training}, we thoroughly select the training datasets where the scale of the pointmaps can be accurately estimated. Specifically, we train our model on three large-scale datasets: ScanNet~\cite{dai2017scannet}, ScanNet++~\cite{yeshwanthliu2023scannetpp} and ARKitScenes~\cite{baruch1arkitscenes}. These datasets primarily comprise of diverse indoor scenes, annotated with ground-truth metric depth and camera poses. We utilize all the data from the official training split of these datasets for training, including a total of 5,929 scenes. For evaluation, we select 10 scenes from the validation split of ScanNet++, and 10 scenes from the validation split of ARKitScenes, which cover various kinds of scenarios. For each evaluation scene, we randomly choose a starting frame from the original image sequence. The lengths of the sampled evaluation frame sequences are set to 2, 10, and 50, with frames sampled at intervals of 5, 3, and 2, respectively. All frames within these sampling intervals are used as ground-truth images for novel-view synthesis.

\vspace{-2mm}

\paragraph{Training details.}
We configure our experimental setup with the number of input views $N_\text{train}=5$, and the number of extra views for photometric supervision {$N_\text{extra}=2$}. We sample the adjunct frames within the interval of $T_\text{min}=5, T_\text{max}=10$. The threshold for masking the photometric loss $\mathcal{L}_\text{MMSE}$ is set to $th_{\alpha} = 1\times 10^{-3}$, and the weight of $\mathcal{L}_\text{MMSE}$ is set to $\lambda=0.1$. The regularization hyperparameter is set to $\alpha=0.4$. We initialize our model with the pretrained weights of Spann3R~\cite{wang20243dreconstructionspatialmemory}, featuring a ViT-large~\cite{dosovitskiy2020image} encoder, ViT-base decoders, a DPT head~\cite{ranftl2021vision} and a 6-layer ViT-based memory network. We train our model at a resolution of $224 \times 224$ on the whole dataset for 8 epochs, with 10,000 samples from each training set per epoch. The training procedure is conducted on 4 H100 GPUs, with a batch size of 8. We employ the AdamW optimizer with a learning rate of $1 \times 10^{-5}$ and a weight decay of $0.05$. During evaluation phase, we apply a confidence threshold of $th_\text{conf}=1.0$ for better rendering quality.

\vspace{-2mm}

\paragraph{Baselines.}
To the best of our knowledge, \ours{} is the first pose-free feed-forward novel-view rendering approach that generalizes to variable-length input views. To effectively evaluate our method, we construct several baselines based on previous works. We compare our model with Spann3R~\cite{wang20243dreconstructionspatialmemory} by projecting the predicted colored pointmaps back onto image planes. Additionally, we evaluate our model against  MVSplat~\cite{chen2024mvsplat}, a feed-forward Gaussian model that requires camera poses; for this comparison, we test MVSplat using ground-truth camera poses. We also include a comparison with the pose-free, optimization-based model, InstantSplat~\cite{fan2024instantsplatsparseviewsfmfreegaussian}. We conduct further comparisons with Splatt3R~\cite{keetha2024splatam} in the 2-view input setting. Since the official Splatt3R model is trained at a resolution of $512 \times 512$, we retrained it on ScanNet++ at a resolution of $224 \times 224$ to ensure a fair comparison. Besides the rendering results, we highlight the efficiency comparison among all the baseline models and \ours{}.

\paragraph{Evaluation metrics.}
We evaluate rendering quality using commonly employed metrics for novel-view synthesis: PSNR, LPIPS, and SSIM. Additionally, we report the overall inference time for all models to compare their efficiency.

\begin{figure*}
    \centering
    \includegraphics[width= \textwidth]{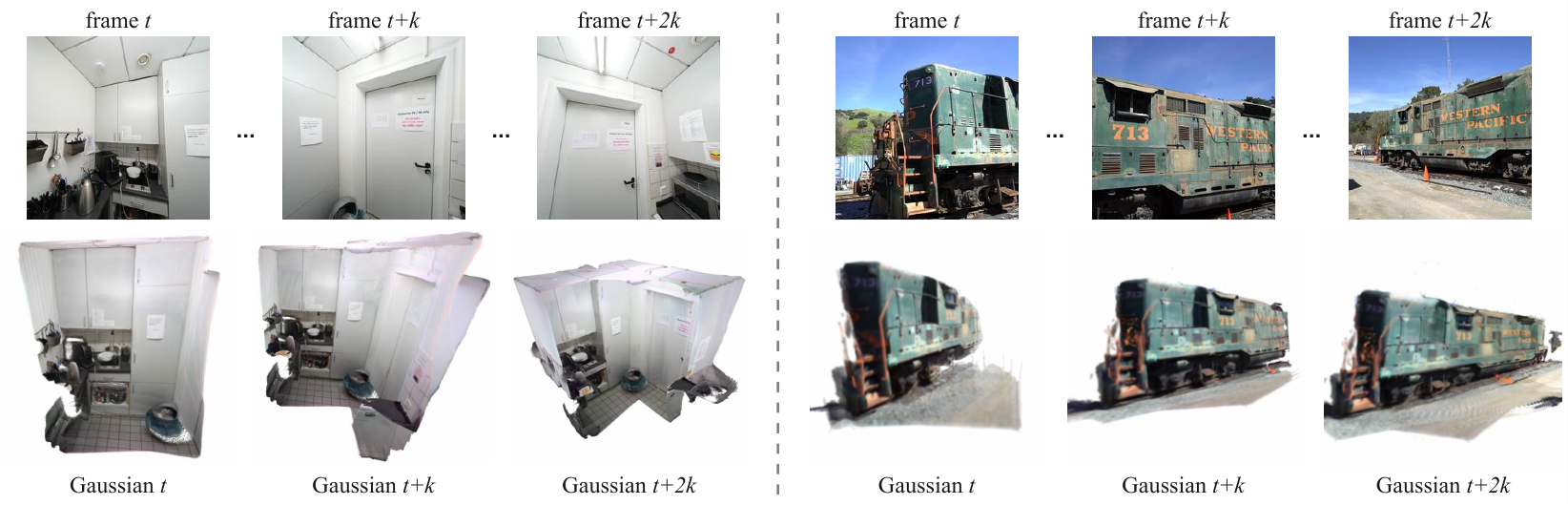}
    \caption{\textbf{\ours{} performs incremental Gaussian reconstruction in real-time.} \textbf{Left}: in-domain scene reconstruction from ScanNet++~\cite{yeshwanthliu2023scannetpp}; \textbf{Right}: out-of-domain scene reconstruction from Tanks and Temples~\cite{Knapitsch2017}.}
    \label{fig:gaussian_vis}
\end{figure*}

\subsection{Experimental results and analysis}

\paragraph{Novel-view rendering quality comparison.}
As illustrated in Tab.~\ref{tab:scannetpp} and Tab.~\ref{tab:arkit}, \ours{} delivers competitive rendering quality as a pose-free, feed-forward model. For pairwise input, \ours{} surpasses most baseline methods across both evaluation datasets, including the optimization-based model IntantSplat~\cite{fan2024instantsplatsparseviewsfmfreegaussian} and the other pose-free feed-forward model Splatt3R~\cite{keetha2024splatam}. However, MVSplat~\cite{chen2024mvsplat}, which relies on ground-truth camera poses, achieves an average of 1.34 dB higher PSNR over our model. In scenarios with 10-view input, \ours{} outperforms all baseline models across all metrics, surpassing the second-best model by 0.74 dB PSNR on ScanNet++ and 1.95 dB on ARKitScenes. Although Splatt3R~\cite{keetha2024splatam} shares the same settings as \ours{}, it is limited to pairwise images as input. In the 50-view input experiment, which poses the greatest challenge, MVSplat encounters CUDA out-of-memory issues and fails to deliver results. In contrast, \ours{} demonstrates strong generalization capabilities, achieving over 20 PSNR on ScanNet++. This highlights \ours's potential in reconstructing Gaussian field from a unlimited image sequence.  Qualitative comparisons are provided in Fig.~\ref{fig:compare_viz}, with scenes sourced from ScanNet++~\cite{yeshwanthliu2023scannetpp} and ARKitScenes~\cite{baruch1arkitscenes}. \ours{} delivers competitive rendering quality and better structural accuracy compared with other methods. Additional details on baseline configurations and result comparisons are provided in the supplementary materials.

\paragraph{Run-time comparison.}
We compare the average running time of \ours{} and the baselines on evaluation scenes. All models are running on a single Nvidia H100 GPU. As shown in Tab.~\ref{tab:scannetpp} and Tab.~\ref{tab:arkit}, \ours{} demonstrates efficiency on par with most feed-forward baseline models while offering the capability to process image sequences of unlimited length. \ours{} maintains constant GPU memory usage by reconstructing Gaussian fields incrementally and leveraging the memory mechanism outlined in Sec.~\ref{subsec:structure}.

\input{table/ablation}

\subsection{Ablation studies}

\paragraph{Extra-view supervision.} As discussed in Sec.~\ref{subsec:training}, we sample $N_\text{extra}=2$ additional views for photometric supervision in training phase. The experiment results in Tab.~\ref{tab:ablation}, indicate that omitting these extra views results in a slight performance reduction, a decrease of 0.33dB on PSNR. 

\vspace{-2mm}
\paragraph{Gaussian pruning.}
We propose pruning the predicted Gaussians using a confidence threshold $th_\text{conf}$ during inference. As demonstrated in Tab.~\ref{tab:ablation}, this approach generally enhances rendering performance. 

We found that while it may result in hollows on particular non-Lambertian surfaces (such as mirrors and windows), Gaussian pruning still helps to eliminate floaters in most of the scenes.

\vspace{-2mm}

\paragraph{Finetuning the backbone.}
We investigate the necessity of finetuning the pretrained backbone of our model. As shown in Tab.~\ref{tab:ablation}, finetuning the backbone leads to slight improvements, by reducing the regression loss $\mathcal{L}_\text{conf}$ and enabling better structure prediction. 
We notice that finetuning the backbone network potentially leads to an increase in regression loss when training is extended over a prolonged number of epochs (e.g., 20 or more). 
This is possibly because in certain scenes where the scene scale is inaccurately estimated, the photometric loss can become significantly large, causing the structural model to degrade.

\vspace{-2mm}

\paragraph{Masked MSE loss.} 
We demonstrate the effectiveness of using a masked MSE loss, as detailed in  Sec.~\ref{subsec:training}. As shown in Tab.~\ref{tab:ablation},  omitting the mask results in a significant performance drop, where the Gaussian Head tends to predict larger scales, leading to blurred rendering results.

\subsection{Discussion}
\label{subsec:discussion}

While \ours{} exhibits competitive efficiency and rendering quality across diverse datasets, it still has several limitations. First, \ours{} is trained on image frames with a relatively high degree of overlap compared to previous works, which causes it to struggle with inputs where adjacent frames have large baselines. Specifically, if there is minimal overlap between $I_t$ and $I_{t+1}$, the reconstruction quality of all subsequent frames following $I_{t+1}$ can deteriorate. Second, since all our training datasets consist of mostly single-room scenes, there may be challenges in generalizing to complex outdoor environments or multi-room architecture. Additionally, we train our model exclusively on a resolution of $224 \times 224$, which may limit its potential to achieve optimal rendering quality.

%% file: table/ablation.tex
\begin{table}
\begin{center}
\setlength{\tabcolsep}{2mm}{
\begin{tabular}{lccc} 
\hline
Method              & PSNR$\uparrow$ & SSIM$\uparrow$ & LPIPS$\downarrow$  \\ 
\hline
w/o extra views       & 22.27 & 0.788 &  0.131      \\
w/o Gaussian pruning    & 22.20 & 0.787 & \textbf{0.128}       \\
w/o finetune backbone & 22.56 & 0.790 &  0.129      \\
w/o loss mask         & 19.70 & 0.619 &  0.288      \\ 
\hline
\textbf{ours}         & \textbf{22.60} & \textbf{0.793} & \textbf{0.128}       \\
\hline
\end{tabular}

\caption{\textbf{Ablation study on 10-view validation scenes from Scannet++.} We show that each component in our proposed method contributes to a better model performance metrics. }
\label{tab:ablation}
}
\end{center}
\vspace{-0.3cm}

\end{table}

%% file: sec/5_conclusion.tex
\section{Conclusion}
\label{sec:conclusion}

We introduce \ours{}, the first pose-free, feed-forward 3D reconstruction pipeline capable of 20~FPS online reconstruction and 200~FPS novel-view synthesis from variable-length sequences of unposed images. Building on the pretrained capabilities of DUSt3R for pairwise 3D structure prediction and leveraging the spatial memory network, \ours{} predicts a robust 3D Gaussian field within a canonical coordinate space, without the need for optimization-based alignment. The Gaussian parameter prediction head with differentiable rasterization enables high-fidelity, photorealistic novel-view synthesis in a purely feed-forward manner. Extensive evaluations demonstrate the effectiveness of \ours{} in both rendering quality and computational efficiency.

%% file: sec/X_suppl.tex
\clearpage
\setcounter{page}{1}
\maketitlesupplementary

\begin{figure}
    \centering
    \includegraphics[width=1\linewidth]{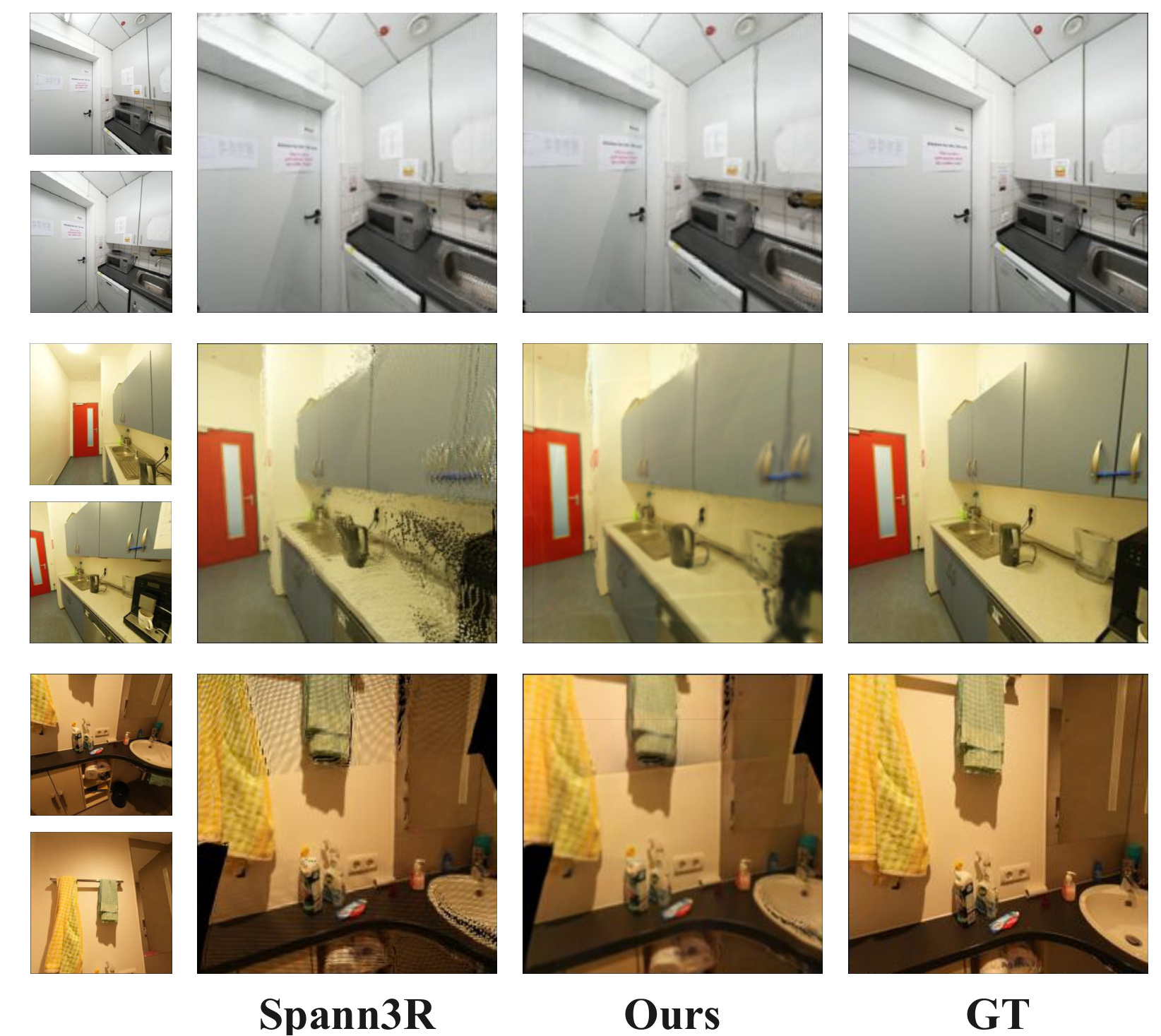}
    \caption{\textbf{Qualitative comparison of novel-view synthesis performance.} \textbf{Top Row}: When the input views are sufficiently dense, Spann3R produces high-quality projected images comparable to images rendered by \ours{}. \textbf{Bottom Two Rows}: In some cases, using colored pointmaps without Gaussian parameters can result in black areas and noticeable floaters when projected back onto image planes. }
    \label{fig:compare_vis_spann3r}
\vspace{-0.3cm}
\end{figure}

\section{More implementation details}
% \label{sec:implementation}

\input{table/supp_Scannetpp}
\input{table/supp_ArkitScenes}

\begin{figure*}[h]
    \centering
    \includegraphics[width= 0.85\textwidth]{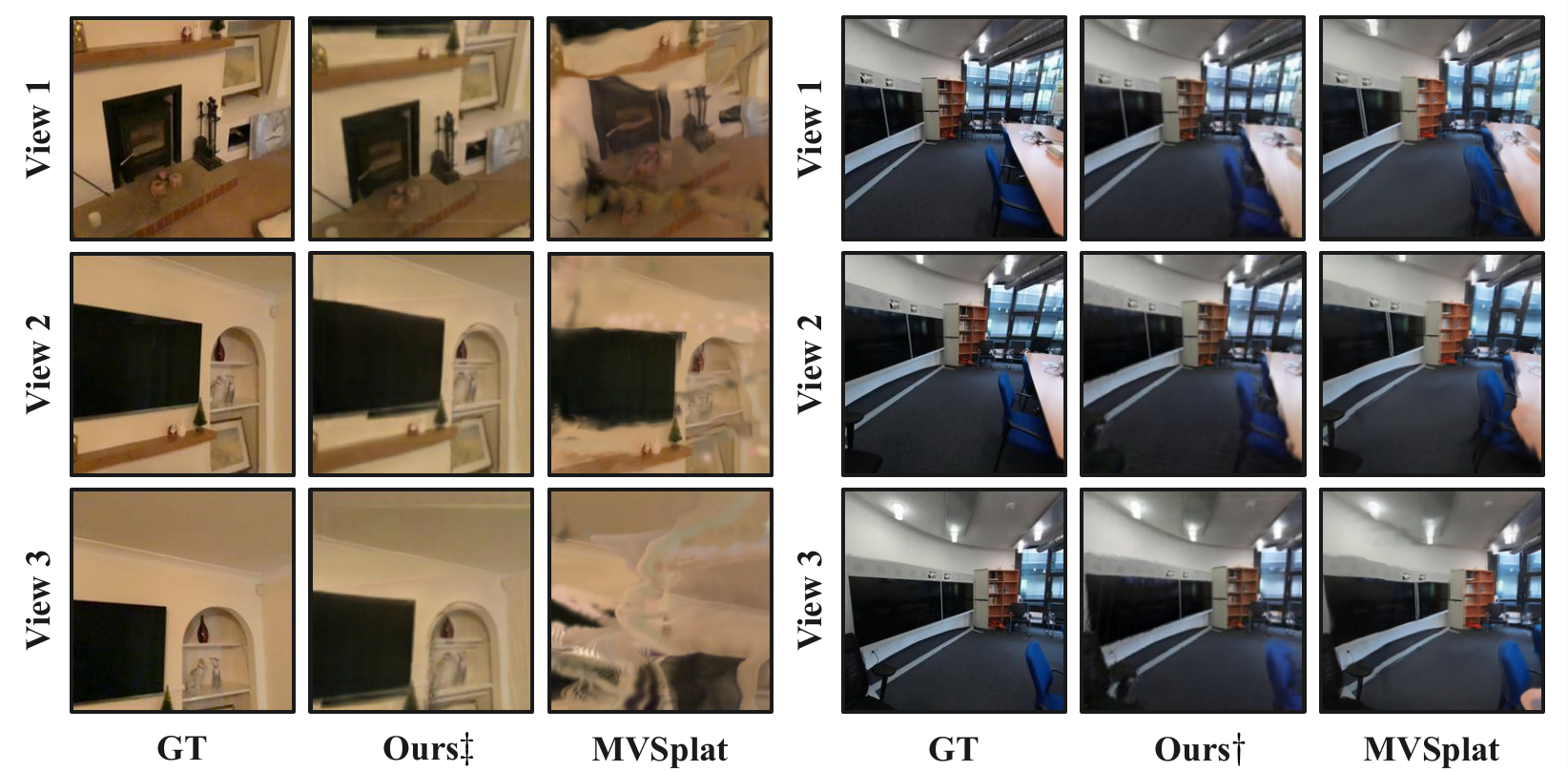}
    \caption{\textbf{Qualitative comparison of cross-dataset novel-view synthesis performance.} \textbf{Left}: visualization of 10-view scenes reconstruction from ARKitScenes~\cite{baruch1arkitscenes}; \textbf{Ours}$\ddag$ represents \ours{} trained on datasets excluding ARKitScenes. \textbf{Right}: visualization of 10-view scenes reconstruction from ScanNet++~\cite{yeshwanthliu2023scannetpp};  \textbf{Ours}$\dag$ represents \ours{} trained on datasets excluding ScanNet++. MVSplat~\cite{chen2024mvsplat} relies on ground-truth camera poses.}
    \label{fig:compare_vis}
\vspace{-0.3cm}
\clearpage
\end{figure*}

As described in Sec.~4.1, we construct several baselines from existing methods. Splatt3R~\cite{keetha2024splatam} is a pose-free feed-forward Gaussian model taking pairwise images as input. The original model is trained on both training and validation splits of ScanNet++~\cite{yeshwanthliu2023scannetpp} at the resolution of $512 \times 512$. To ensure fair comparisons, we retrain the model on the training split of ScanNet++ at the resolution of $224 \times 224$. We keep the same training parameters used in original paper. We attempt to include ARKitScenes~\cite{baruch1arkitscenes} and ScanNet~\cite{dai2017scannet} in the training set as we used to train \ours{}, but the quality of the rendered output deteriorated significantly within the first epoch. 

For pose-required model MVSplat~\cite{chen2024mvsplat}, we evaluate both pretrained model weights trained on RE10K~\cite{zhou2018stereo} and ACID~\cite{infinite_nature_2020} respectively. Considering that our evaluation sets mainly comprise indoor scenes, which are more similar to those in RE10K, the model trained on RE10K significantly outperforms the version train on ACID. Since MVSplat uses a cross-dataset evaluation, for fair comparisons, we additionally train \ours{} on ScanNet and ARKitScenes, and then evaluate it on ScanNet++. Equivalently, we also train \ours{} on ScanNet and ScanNet++ and evaluate it on ARKitScenes. The performance comparisons are presented in Tab.~\ref{tab:supp_arkit} and Tab.~\ref{tab:supp_scannetpp}, and are further analyzed in Sec.~\ref{sec:more_experiment}.

Furthermore, for optimization-based InstantSplat~\cite{fan2024instantsplatsparseviewsfmfreegaussian}, we adopt the setting of \textbf{InstantSplat-S} model in the original paper, with $200$ iterations for Gaussian optimization, and $500$ iterations per-image for test-time optimization (TTO). There is no doubt that increasing the optimization steps, as with \textbf{InstantSplat-XL} in the original InstantSplat paper, would enhance the rendering quality. However, the \textbf{InstantSplat-S} is already $100\times$ slower than our model, and our aim is not to outperform all the heavily optimized models.

\section{More experimental results}
\label{sec:more_experiment}

We provide a qualitative comparison with Spann3R~\cite{wang20243dreconstructionspatialmemory} in Fig.~\ref{fig:compare_vis_spann3r}. Although we share a similar structural model with Spann3R, the rendering results of \ours{} surpasses Spann3R significantly. This is because using colored pointmaps without Gaussian parameters can result in black areas and noticeable floaters when projected back onto image planes. However, when the input views are sufficiently dense, meaning there is substantial overlap between adjacent frames, Spann3R produces high-quality projected images comparable to those rendered by \ours{}.

As shown in Tab.~\ref{tab:supp_scannetpp} and Tab.~\ref{tab:supp_arkit}, \textbf{ours}$\dag$ refers 
to \ours{} trained on the training sets excluding ScanNet++ and \textbf{ours}$\ddag$ denotes \ours{} trained on training sets excluding ARKitScenes. It is evident that \ours{} is less effective on 2-view input scenes but outperforms pose-dependent MVSplat on 10-view input scenes during cross-dataset evaluation on both ScanNet++ and ARKitScenes datasets. In most cross-dataset test cases, \ours{} exhibits slightly diminished detail and somewhat reduced color and structural accuracy compared to when it is trained on the full dataset. A qualitative comparison is demonstrated in Fig.~\ref{fig:compare_vis}.

\section{Limitations and future works}
\label{sec:limitations}

Through empirical experiments, \ours{} demonstrates competitive efficiency and rendering quality across a variety of datasets, but it retains several limitations. First, \ours{} is trained on image frames with dense input views, with substantial overlap between adjacent frames. In contrast, previous works often effectively manage inputs with larger baselines. In practical scenarios, if the overlap between input frames $I_t$ and $I_{t+1}$ is minimal, the reconstruction quality of all subsequent frames following $I_{t+1}$ will deteriorate. Second, since all our training datasets primarily consist of single-room scenes, there may be challenges in generalizing to complex outdoor environments or multi-room architecture. Additionally, we train our model exclusively at the resolution of $224 \times 224$, which may limit its potential to achieve optimal rendering quality.

Another promising direction for future work is to incorporate an optimization-based approach for camera pose estimation from the predicted Gaussians, similar to the methods used in DUSt3R\cite{wang2024DUSt3R} and MASt3R\cite{leroy2024groundingimagematching3d}. These methodologies leverage advanced optimization techniques to refine camera pose estimates, potentially enhancing the accuracy and robustness of the reconstructed 3D models. By integrating such optimization strategies, we could address current limitations related to pose estimation in sparse or complex scene environments. Moreover, this enhancement could support the extension of \ours{} to a wider range of applications.

%% file: table/supp_Scannetpp.tex
\begin{table*}[ht]
\begin{center}
\setlength{\tabcolsep}{0.8mm}{

\begin{tabular}{llcccccccccccc} 
\hline
\multirow{2}{*}{PF} & \multirow{2}{*}{Method} & \multicolumn{4}{c}{2 views}                                       & \multicolumn{4}{c}{10 views}                                      & \multicolumn{4}{c}{50 views}                                       \\ 
\cline{3-14}
             & & PSNR$\uparrow$           & SSIM$\uparrow$           & LPIPS$\downarrow$         & time           & PSNR$\uparrow$          & SSIM$\uparrow$          & LPIPS$\downarrow$         & time           & PSNR$\uparrow$          & SSIM $\uparrow$          & LPIPS $\downarrow$         & time            \\ 
\hline
  &   {MVSplat (ACID)}                 &  {19.75}            & {0.662}         &    {0.250}      &   {0.083}             &   {15.21}             &  {0.518}              &    {0.458}            &      {0.579}           &      {-}          &      {-}          &     {-}           &       {-}          \\
  &   {MVSplat (RE10K)}                 &  \textbf{23.67}            & \textbf{0.811}         &    \textbf{0.180}      &   {0.081}             &   {19.31}             &  {0.697}              &    {0.334}            &      {0.554}           &      {-}          &      {-}          &     {-}           &       {-}          \\

\checkmark          &  \textbf{ours}$\dag$           & {21.52} & {0.785} & {0.153} & \textbf{0.138} & \textbf{20.70} & \textbf{0.739} & \textbf{0.198} & \textbf{0.512} & \textbf{19.79} & \textbf{0.676} & \textbf{0.265} & \textbf{2.179}  \\
\hline
\end{tabular}

\caption{\textbf{Cross-dataset novel-view synthesis performance on ScanNet++~\cite{yeshwanthliu2023scannetpp}}. All metrics are averaged over 10 validation scenes. `PF' denotes Pose-Free. Average running time is measured in seconds. MVSplat~\cite{chen2024mvsplat} encounters CUDA out-of-memory on 50-view scenes on H100 GPU. \textbf{ous}$\dag$ represents \ours{} trained on original training sets excluding ScanNet++.}
\label{tab:supp_scannetpp}

}
\end{center}
\vspace{-0.5cm}
\end{table*}

%% file: table/supp_ArkitScenes.tex
\begin{table*}[ht]
\begin{center}
\setlength{\tabcolsep}{0.8mm}{

\begin{tabular}{llcccccccccccc} 
\hline
\multirow{2}{*}{PF} & \multirow{2}{*}{Method} & \multicolumn{4}{c}{2 views}                                       & \multicolumn{4}{c}{10 views}                                      & \multicolumn{4}{c}{50 views}                                       \\ 
\cline{3-14}
    &                         & PSNR$\uparrow$           & SSIM$\uparrow$           & LPIPS$\downarrow$         & time           & PSNR$\uparrow$          & SSIM$\uparrow$          & LPIPS$\downarrow$         & time           & PSNR$\uparrow$          & SSIM $\uparrow$          & LPIPS $\downarrow$         & time            \\ 
\hline
  &   {MVSplat (ACID)}                 &  {18.65}            & {0.594}         &    {0.307}      &   0.089             &   {15.00}             &  {0.508}              &    {0.482}            &      {0.537}           &      {-}          &      {-}          &     {-}           &       {-}          \\
        & {MVSplat (RE10K)}                 &    \textbf{23.00}            &     \textbf{0.782}           &    {0.210}            &    \textbf{0.079}            &    {19.43}           &    \textbf{0.694}            &     {0.357}           &      \textbf{0.506}          &     {-}           &          {-}      &     {-}           &    {-}             \\

\checkmark          & \textbf{ours}$\ddag$           & {19.98} & {0.691} & \textbf{0.209} & {0.149}        & \textbf{20.47} & {0.665} & \textbf{0.274} & {0.532} & \textbf{17.84} & \textbf{0.507} & \textbf{0.367} & \textbf{2.511}  \\
\hline
\end{tabular}
\caption{\textbf{Cross-dataset novel-view synthesis performance on ARKitScenes~\cite{baruch1arkitscenes}}. All metrics are averaged over 10 validation scenes. Average running time is measured in seconds. Experimental settings and evaluation criteria are consistent with those described in Tab.~\ref{tab:scannetpp}. \textbf{ous}$\ddag$ represents \ours{} trained on original training sets excluding ARKitScenes.}
\label{tab:supp_arkit}
}
\end{center}
\vspace{-0.5cm}
\end{table*}